\documentclass[11pt,a4paper]{article}
\usepackage[hyperref]{emnlp2018}
\usepackage{times}
\usepackage{latexsym}

\usepackage{url}

\usepackage{bm}
\usepackage{amsfonts, amsmath, amssymb}
\usepackage{algorithm}
\usepackage{algorithmic}
\usepackage{enumerate}
\usepackage{graphicx}
\usepackage{color}
\usepackage{booktabs}
\usepackage{multirow}
\usepackage{tabularx}
\usepackage{cleveref}
\usepackage{subfigure}
\usepackage[font=small]{caption}
\usepackage{titlesec}
\usepackage{arydshln}
\usepackage{enumitem}
\usepackage{makecell}
\usepackage{pbox}

\crefformat{section}{\S #2#1#3} 
\crefformat{subsection}{\S #2#1#3}
\crefformat{subsubsection}{\S #2#1#3}
\aclfinalcopy 
\def\aclpaperid{336} 

\titlespacing{\paragraph}{%
  0pt}{
  0.3\baselineskip}{
  1em}%

\title{Unsupervised Learning of Syntactic Structure \\ with Invertible Neural Projections}

\author{Junxian He \ \ \
  Graham Neubig \ \ \
  Taylor Berg-Kirkpatrick\\
  Language Technologies Institute \\
  School of Computer Science \\
  Carnegie Mellon University \\
  {\tt \{junxianh, gneubig, tberg\}@cs.cmu.edu}\\}

\date{}

\begin{document}
\maketitle
\begin{abstract}
Unsupervised learning of syntactic structure is typically performed using generative models with discrete latent variables and multinomial parameters. In most cases, these models have not leveraged continuous word representations. In this work, we propose a novel generative model that jointly learns discrete syntactic structure and continuous word representations in an unsupervised fashion by cascading an \textit{invertible} neural network with a structured generative prior. We show that the invertibility condition allows for efficient exact inference and marginal likelihood computation in our model so long as the prior is well-behaved. 
In experiments we instantiate our approach with both Markov and tree-structured priors, evaluating on two tasks: part-of-speech (POS) induction, and unsupervised dependency parsing without gold POS annotation. On the Penn Treebank, our Markov-structured model surpasses state-of-the-art results on POS induction. Similarly, we find that our tree-structured model achieves state-of-the-art performance on unsupervised dependency parsing for the difficult training condition where neither gold POS annotation nor punctuation-based constraints are available.\footnote{Code is available at \href{https://github.com/jxhe/struct-learning-with-flow}{https://github.com/jxhe/struct-learning-with-flow}.}

\end{abstract}

\section{Introduction}

Data annotation is a major bottleneck for the application of supervised learning approaches to many problems. As a result, unsupervised methods that learn directly from unlabeled data are increasingly important. For tasks related to unsupervised syntactic analysis, discrete generative models have dominated in recent years -- for example,
for both part-of-speech (POS) induction ~\citep{blunsom2011hierarchical, stratos2016unsupervised} and unsupervised dependency parsing~\citep{klein2004corpus, cohen2009shared, pate2016grammar}. While similar models have had success on a range of unsupervised tasks, they have mostly ignored the apparent utility of continuous word representations evident from supervised NLP applications~\citep{he2017deep, peters2018deep}. In this work, we focus on leveraging and explicitly representing continuous word embeddings within unsupervised models of syntactic structure.
\begin{figure}[!t]
  \centering
   \subfigure[Traditional pre-trained \newline \hspace*{1.6em}skip-gram embeddings]{\includegraphics[scale=0.18]{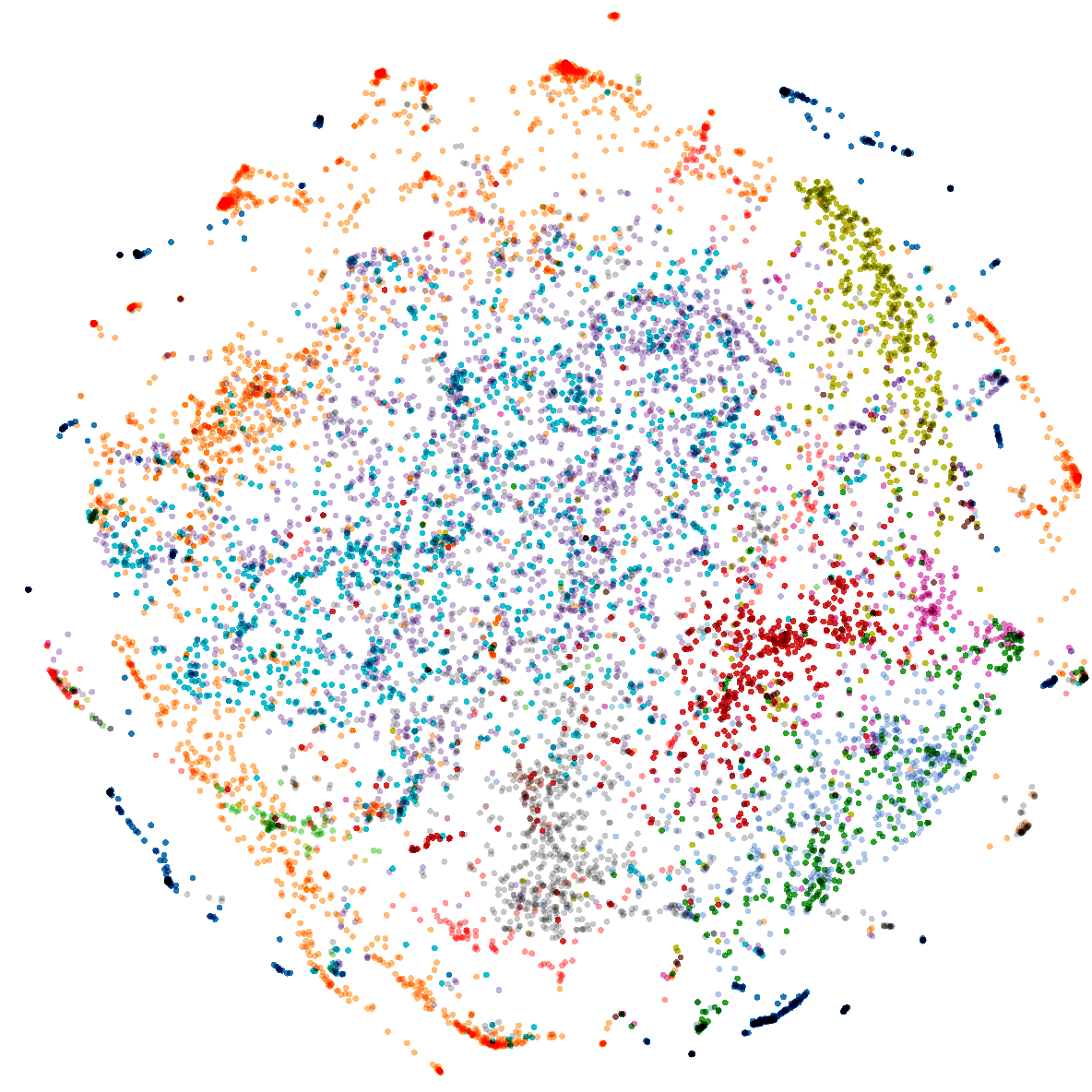}\label{fig:vi-pos-skip}}
    \subfigure[Learned latent embedd- \newline \hspace*{1.6em}ings from our approach]{\includegraphics[scale=0.18]{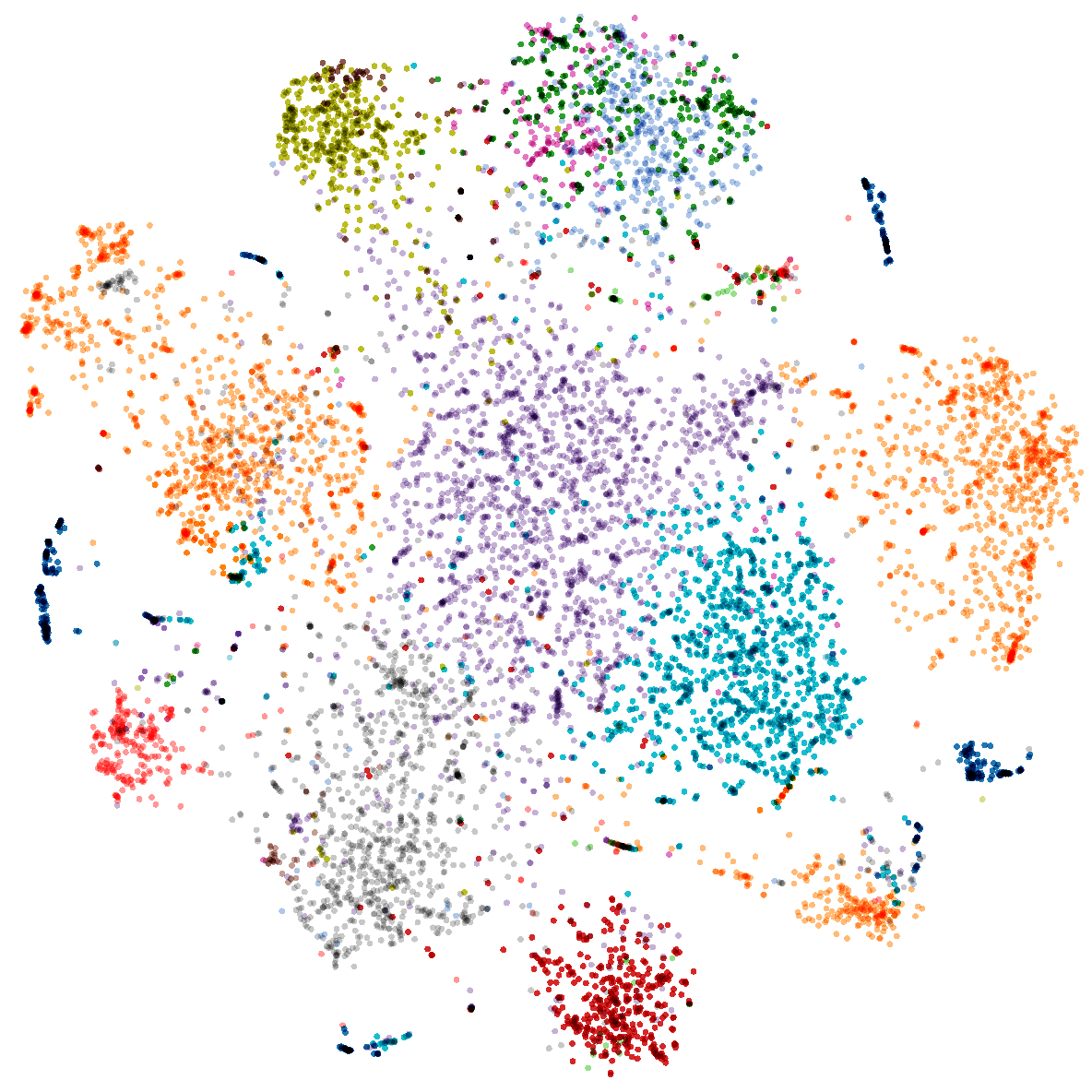}\label{fig:vi-pos-mar}}
\vspace{-2mm}
\caption{Visualization (t-SNE) of skip-gram embeddings (trained on one billion words with context window size equal to 1) and latent embeddings learned by our approach with a Markov-structured prior. Each node represents a word and is colored according to the most likely gold POS tag from the Penn Treebank (best seen in color). }
\label{fig:vi-pos}
\vspace{-5mm}
\end{figure}

\begin{figure*}[!t]
   \centering
    \includegraphics[scale=0.45]{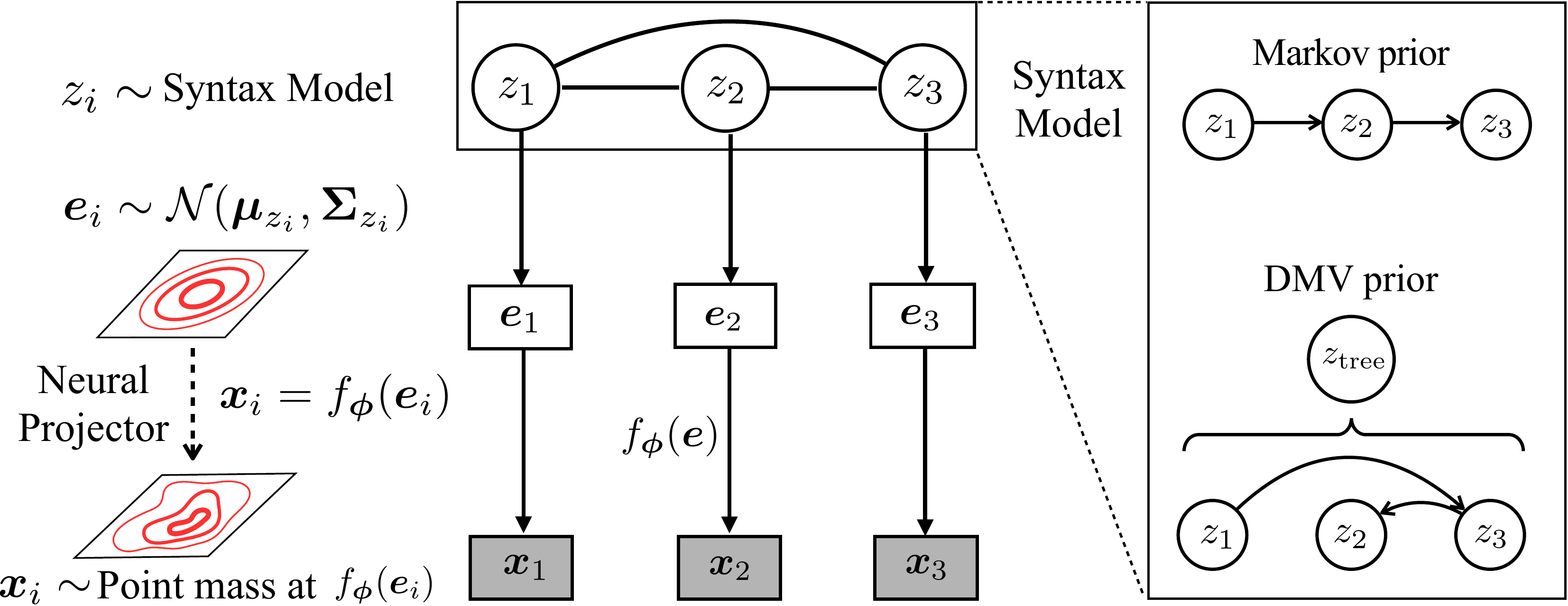}
    \vspace{-1.5mm}
    \caption{Depiction of proposed generative model. The syntax model is composed of discrete random variables, $z_i$. Each $\bm{e}_i$ is a latent continuous embeddings sampled from Gaussian distribution conditioned on $z_i$, while $\bm{x}_i$ is the observed embedding, deterministically derived from $\bm{e}_i$. The left portion depicts how the neural projector maps the simple Gaussian  to a more complex distribution in the output space. The right portion depicts two instantiations of the syntax model in our approach: one is Markov-structured and the other is DMV-structured. For DMV, $z_{\text{tree}}$ is the latent dependency tree structure.}
    \label{fig:model-scheme}
  \vspace{-5mm}
\end{figure*}
Pre-trained word embeddings from massive unlabeled corpora offer a compact way of injecting a prior notion of word similarity into models that would otherwise treat words as discrete, isolated categories. 
However, the specific properties of language captured by any particular embedding scheme can be difficult to control, and, further, may not be ideally suited to the task at hand. For example, pre-trained skip-gram embeddings~\citep{mikolov2013efficient} with small context window size are found to capture the syntactic properties of language well~\citep{bansal2014tailoring, lin2015unsupervised}. However, if our goal is to separate syntactic categories, this embedding space is not ideal -- POS categories correspond to overlapping interspersed regions in the embedding space, evident in Figure~\ref{fig:vi-pos-skip}.

In our approach, we propose to learn a new latent embedding space as a projection of pre-trained embeddings (depicted in Figure~\ref{fig:vi-pos-mar}), while \emph{jointly} learning latent syntactic structure -- for example, POS categories or syntactic dependencies.
To this end, we introduce 
a new generative model  (shown in Figure~\ref{fig:model-scheme}) that first generates a latent syntactic representation (e.g. a dependency parse) from a discrete structured prior (which we also call the ``syntax model''),
then, conditioned on this representation, generates a sequence of latent embedding random variables corresponding to each word, and finally produces the observed (pre-trained) word embeddings by projecting these latent vectors through a parameterized non-linear function.
 The latent embeddings can be jointly learned with the structured syntax model in a completely unsupervised fashion. 

By choosing an invertible neural network as our non-linear projector, and then parameterizing our model in terms of the projection's inverse, we are able to derive tractable exact inference and marginal likelihood computation procedures so long as inference is tractable in the underlying syntax model.  In \cref{sec:learn-with-inv} we show that this derivation corresponds to an alternate view of our approach whereby we jointly learn a mapping of observed word embeddings to a new embedding space that is more suitable for the syntax model, but include an additional Jacobian regularization term to prevent information loss. 

Recent work has sought to take advantage of word embeddings in unsupervised generative models with alternate approaches~\citep{lin2015unsupervised, tran2016unsupervised, jiang2016unsupervised, han2017dependency}. \citet{lin2015unsupervised} build an HMM with Gaussian emissions on observed word embeddings, but they do not attempt to learn new embeddings.  \citet{tran2016unsupervised}, \citet{jiang2016unsupervised}, and \citet{han2017dependency} extend HMM or dependency model with valence (DMV)~\citep{klein2004corpus} with multinomials that use word (or tag) embeddings in their parameterization. However, they do not represent the embeddings as latent variables. 

In experiments, we instantiate our approach using both a Markov-structured syntax model and a tree-structured syntax model -- specifically, the DMV. We evaluate on two tasks: part-of-speech (POS) induction and unsupervised dependency parsing without gold POS tags. Experimental results on the Penn Treebank~\citep{marcus1993building} demonstrate that our approach improves the basic HMM and DMV by a large margin, leading to the state-of-the-art results on POS induction, and state-of-the-art results on unsupervised dependency parsing in the difficult training scenario where neither gold POS annotation nor punctuation-based constraints are available. 

%

\vspace{-0.4em}
\section{Model}
\vspace{-0.4em}
\label{sec:model}
As an illustrative example, we first present a baseline model for Markov syntactic structure (POS induction) that treats a sequence of pre-trained word embeddings as observations. Then, we propose our novel approach, again using Markov structure, that introduces latent word embedding variables and a neural projector. Lastly, we extend our approach to more general syntactic structures.

\subsection{Example: Gaussian HMM}

We start by describing the Gaussian hidden Markov model introduced by~\citet{lin2015unsupervised}, which is a locally normalized model with multinomial transitions and Gaussian emissions. Given a sentence of length $\ell$, we denote the latent POS tags as $\bm{z} = \{z_i\}_{i=1}^\ell$, observed (pre-trained) word embeddings as $\bm{x} = \{\bm{x}_i\}_{i=1}^\ell$, transition parameters as $\bm{\theta}$, and Gaussian emission parameters as $\bm{\eta}$. The joint distribution of data and latent variables factors as:
\vspace{-3mm}
\begin{equation}
\hspace{-0.26mm}
p(\bm{z}, \bm{x}; \bm{\theta}, \bm{\eta}) = \prod\nolimits_{i=1}^\ell p_{\bm{\theta}}(z_i | z_{i-1}) p_{\bm{\eta}}(\bm{x}_i | z_i),
\end{equation}
where $p_{\bm{\theta}}(z_i | z_{i-1})$ is the multinomial transition probability and $p_{\bm{\eta}}(\bm{x}_i | z_i)$ is the multivariate Gaussian emission probability. 



While the observed word embeddings do inform this model with a notion of word similarity -- lacking in the basic multinomial HMM -- the Gaussian emissions may not be sufficiently flexible to separate some syntactic categories in the complex pre-trained embedding space -- for example the skip-gram embedding space as visualized in Figure~\ref{fig:vi-pos-skip} where different POS categories overlap. 
Next we introduce a new approach that adds flexibility to the emission distribution by incorporating new latent embedding variables.

\subsection{Markov Structure with Neural Projector}
\label{sec:hmm-neural}
To flexibly model observed embeddings and yield a new representation space that is more suitable for the syntax model, we propose to cascade a neural network as a projection function, deterministically transforming the simple space defined by the Gaussian HMM to the observed embedding space. We denote the latent embedding of the $i^{th}$ word in a sentence as $\bm{e}_i\in \mathbb{R}^{d_e}$, and the neural projection function as $f$, parameterized by $\bm{\phi}$. In the case of sequential Markov structure, our new model corresponds to the following generative process:

\vspace{2mm}
\noindent For each time step $i = 1, 2, \cdots, \ell$,
\vspace{-2mm}
  \begin{itemize}[noitemsep, leftmargin=*]
   \item Draw the latent state ${z_i} \sim p_{\bm{\theta}}(z_i | z_{i-1})$ 
   \item Draw the latent embedding $\bm{e}_i \sim \mathcal{N}(\bm{\mu}_{z_i}, \bm{\Sigma}_{z_i})$
   \item Deterministically produce embedding \\$\bm{x}_i\ = f_{\bm{\phi}}(\bm{e}_i)$
  \end{itemize}
\vspace{0mm}

\noindent The graphical model is depicted in Figure~\ref{fig:model-scheme}. The deterministic projection can also be viewed as sampling each observation from a point mass at $f_{\bm{\phi}}(\bm{e}_i)$.
The joint distribution of our model is:
\begin{equation}
\label{eq:joint-prob}
\begin{split}
& p(\bm{z}, \bm{e}, \bm{x}; \bm{\theta}, \bm{\eta}, \bm{\phi}) \\
& = \prod\nolimits_{i=1}^\ell [p_{\bm{\theta}}(z_i | z_{i-1})p_{\bm{\eta}}(\bm{e}_i | z_i) p_{\bm{\phi}}(\bm{x}_i | \bm{e}_i)],
\end{split}
\end{equation}
where $p_{\bm{\eta}}(\cdot | z_i)$ is a conditional Gaussian distribution, and $p_{\bm{\phi}}(\bm{x}_i | \bm{e}_i)$ is the Dirac delta function centered at $f_{\bm{\phi}}(\bm{e}_i)$:
\begin{equation}
\label{eq:dirac}
p_{\bm{\phi}}(\bm{x}_i | \bm{e}_i) = \delta(\bm{x}_i-f_{\bm{\phi}}(\bm{e}_i)) = 
\begin{cases}
\infty & \hspace{-2mm}\bm{x}_i = f_{\bm{\phi}}(\bm{e}_i)\\
0 & \hspace{-2mm}\text{otherwise}
\end{cases}
\raisetag{10pt}
\end{equation}

\subsection{General Structure with Neural Projector}
\label{sec:general-neural}
Our approach can be applied to a broad family of structured syntax models. We denote latent embedding variables as $\bm{e}=\{\bm{e}_i\}_{i=1}^{\ell}$, discrete latent variables in the syntax model as $\bm{z}=\{z_k\}_{k=1}^K$ ($K \geqslant \ell$), where $z_1, z_2, \ldots, z_{\ell}$ are conditioned to generate $\bm{e}_1, \bm{e}_2, \ldots, \bm{e}_{\ell}$. The joint probability of our model factors as:
\vspace{-1mm}
\begin{equation}
\label{eq:prob-general}
\begin{split}
p(\bm{z}, \bm{e}, \bm{x}; \bm{\theta}, \bm{\eta}, \bm{\phi}) =& \prod\nolimits_{i=1}^{\ell}\big[p_{\bm{\eta}}(\bm{e}_i | z_i)p_{\bm{\phi}}(\bm{x}_i | \bm{e}_i)\big] \\
& \cdot p_{\text{syntax}}(\bm{z}; \bm{\theta}),
\end{split}
\raisetag{0.9\baselineskip}
\end{equation}
where $p_{\text{syntax}}(\bm{z}; \bm{\theta})$ represents the probability of the syntax model, and can encode any syntactic structure -- though, its factorization structure will determine whether inference is tractable in our full model.
As shown in Figure~\ref{fig:model-scheme}, we focus on two syntax models for syntactic analysis in this paper. The first is Markov-structured, which we use for POS induction, and the second is DMV-structured, which we use to learn dependency parses without supervision. 

The marginal data likelihood of our model is:
\begin{equation}
\label{eq:margin-obj}
\begin{split}
p(\bm{x}) = &\sum\nolimits_{\bm{z}}\Big(p_{\text{syntax}}(\bm{z}; \bm{\theta}) \\
&\cdot  \prod\nolimits_{i=1}^{\ell}\big[\underbrace{\int_{\bm{e}_i}p_{\bm{\eta}}(\bm{e}_i | z_i)p_{\bm{\phi}}(\bm{x}_i | \bm{e}_i)\text{d}\bm{e}_i}_{p(\bm{x}_i | z_i)}\big]\Big).
\end{split}
\end{equation}
While the discrete variables $\bm{z}$ can be marginalized out with dynamic program in many cases, it is generally intractable to marginalize out the latent continuous variables, $\bm{e}_i$, for an arbitrary projection $f$ in Eq.~\eqref{eq:margin-obj}, which means inference and learning may be difficult.
In \cref{sec:opt}, we address this issue by constraining $f$ to be invertible, and show that this constraint enables tractable exact inference and marginal likelihood computation.

\section{Learning \& Inference}

\label{sec:opt}


In this section, we introduce an invertibility condition for our neural projector to tackle the optimization challenge. Specifically, we constrain our neural projector with two requirements: (1) $\text{dim}(\bm{x}) = \text{dim}(\bm{e})$ and (2) $f_{\bm{\phi}}^{-1}$ exists. Invertible transformations have been explored before in independent components analysis~\citep{hyvarinen2004independent}, gaussianization~\citep{chen2001gaussianization}, and deep density models~\citep{dinh2014nice, dinh2016density, kingma2018glow}, for unstructured data. Here, we generalize this style of approach to \emph{structured} learning, and augment it with discrete latent variables ($z_i$). Under the invertibility condition, we derive a learning algorithm and give another view of our approach revealed by the objective function. Then, we present the architecture of a neural projector we use in experiments: a volume-preserving invertible neural network proposed by \citet{dinh2014nice} for independent components estimation.


\subsection{Learning with Invertibility}
\label{sec:learn-with-inv}
For ease of exposition, we explain the learning algorithm in terms of Markov structure without loss of generality. As shown in Eq.~\eqref{eq:margin-obj}, the optimization challenge in our approach comes from the intractability of the marginalized emission factor $p(\bm{x}_i|z_i)$. If we can marginalize out $\bm{e}_i$ and compute $p(\bm{x}_i|z_i)$, 
then the posterior and marginal likelihood of our Markov-structured model can be computed with the forward-backward algorithm.  We can apply Eq.~\eqref{eq:dirac} and obtain : 
\begin{equation*}
p(\bm{x}_i | z_i; \bm{\eta}, \bm{\phi}) = \int_{\bm{e}_i}p_{\bm{\eta}}(\bm{e}_i|z_i)\delta(\bm{x}_i-f_{\bm{\phi}}(\bm{e}_i))\text{d}\bm{e}_i.
\end{equation*}
By using the change of variable rule to the integration, which allows the integration variable $\bm{e}_i$ to be replaced by $\bm{x}_i^{\prime}=f_{\bm{\phi}}(\bm{e}_i)$, the marginal emission factor can be computed in closed-form when the invertibility condition is satisfied:
\begin{equation}
\label{eq:emission}
\begin{split}
&p(\bm{x}_i | \bm{z}_i; \bm{\eta}, \bm{\phi}) \\
&= \int_{\bm{x}_i^{\prime}}p_{\bm{\eta}}(f^{-1}_{\bm{\phi}}(\bm{x}_i^{\prime})|z_i)\delta(\bm{x}_i-\bm{x}_i^{\prime})\Big|\text{det} \frac{\partial f^{-1}_{\bm{\phi}}}{\partial \bm{x}_i^{\prime}}\Big|\text{d}\bm{x}_i^{\prime}\\
&= p_{\bm{\eta}}(f^{-1}_{\bm{\phi}}(\bm{x}_i)|z_i)\Big|\text{det}\frac{\partial f^{-1}_{\bm{\phi}}}{\partial \bm{x}_i}\Big|,
\end{split}
\raisetag{1.2\baselineskip}
\end{equation}
where $p_{\bm{\eta}}(\cdot | z)$ is a conditional Gaussian distribution, $\frac{\partial f^{-1}_{\bm{\phi}}}{\partial \bm{x}_i}$ is the Jacobian matrix of function $f_{\bm{\phi}}^{-1}$ at $\bm{x}_i$, and $\big|\text{det}\frac{\partial f^{-1}_{\bm{\phi}}}{\partial \bm{x}_i}\big|$ represents the absolute value of its determinant. This Jacobian term is nonzero and differentiable if and only if $f_{\bm{\phi}}^{-1}$ exists.

Eq.~\eqref{eq:emission} shows that we can directly calculate the marginal emission distribution $p(\bm{x}_i|z_i)$.
Denote the marginal data likelihood of Gaussian HMM as $p_{\text{{\tiny HMM}}}(\bm{x})$, then the log marginal data likelihood of our model can be directly written as:
\begin{equation}\setlength\abovedisplayskip{6pt}\setlength\belowdisplayskip{6pt}
\label{eq:margin-ll}
\begin{split}
\log p(\bm{x}) = &\log p_{\text{{\tiny HMM}}}(f^{-1}_{\bm{\phi}}(\bm{x})) \\
&+ \sum\nolimits_{i=1}^{\ell}\log \Big|\text{det} \frac{\partial f^{-1}_{\bm{\phi}}}{\partial \bm{x}_i}\Big|,
\end{split}
\end{equation}
where $f^{-1}_{\bm{\phi}}(\bm{x})$ represents the new sequence of embeddings after applying $f_{\bm{\phi}}^{-1}$ to each $\bm{x}_i$. Eq.~\eqref{eq:margin-ll} shows that the training objective of our model is simply the Gaussian HMM log likelihood with an additional Jacobian regularization term. 
%
From this view, our approach can be seen as equivalent to reversely projecting the data through $f^{-1}_{\bm{\phi}}$ to another manifold $\bm{e}$ that is directly modeled by the Gaussian HMM, with a regularization term. Intuitively, we optimize the reverse projection $f^{-1}_{\bm{\phi}}$ to modify the $\bm{e}$ space, making it more appropriate for the syntax model. The Jacobian regularization term accounts for the volume expansion or contraction behavior of the projection. Maximizing it can be thought of as preventing information loss. In the extreme case, the Jacobian determinant is equal to zero, which means the projection is non-invertible and thus information is being lost through the projection. Such ``information preserving'' regularization is crucial during optimization, otherwise the trivial solution of always projecting data to the same single point to maximize likelihood is viable.\footnote{For example, all $\bm{e}_i$ could learn to be zero vectors, leading to the trivial solution of learning zero mean and zero variance Gaussian emissions achieving infinite data likelihood.}

More generally, for an arbitrary syntax model the data likelihood of our approach is:
\begin{equation}
\label{eq:learn-general}
\begin{split}
\hspace{-3mm}
p(\bm{x}) = &\sum\nolimits_{\bm{z}}\Big(p_{\text{syntax}}(\bm{z}) \\
& \cdot\prod\nolimits_{i=1}^{\ell}p_{\bm{\eta}}(f^{-1}_{\bm{\phi}}(\bm{x}_i) | z_i)\Big|\text{det}\frac{\partial f^{-1}_{\bm{\phi}}}{\partial \bm{x}_i}\Big|\Big).
\end{split}
\end{equation}
If the syntax model itself allows for tractable inference and marginal likelihood computation, the same dynamic program can be used to marginalize out $\bm{z}$. Therefore, our joint model inherits the tractability of the underlying syntax model.

\subsection{Invertible Volume-Preserving Neural Net}
For the projection we can use an arbitrary invertible function, and given the representational power of neural networks they seem a natural choice. However, calculating the inverse and Jacobian of an arbitrary neural network can be difficult, as it requires that all component functions be invertible and also requires storage of large Jacobian matrices, which is memory intensive.
To address this issue, several recent papers propose specially designed invertible networks that are easily trainable yet still powerful~\citep{dinh2014nice, dinh2016density, jacobsen2018revnet}. Inspired by these works, we use the invertible transformation proposed by \citet{dinh2014nice}, which consists of a series of ``coupling layers''.  This architecture is specially designed to guarantee a unit Jacobian determinant (and thus the invertibility property).

\begin{figure}[!t]
\centering
    \includegraphics[scale=0.14]{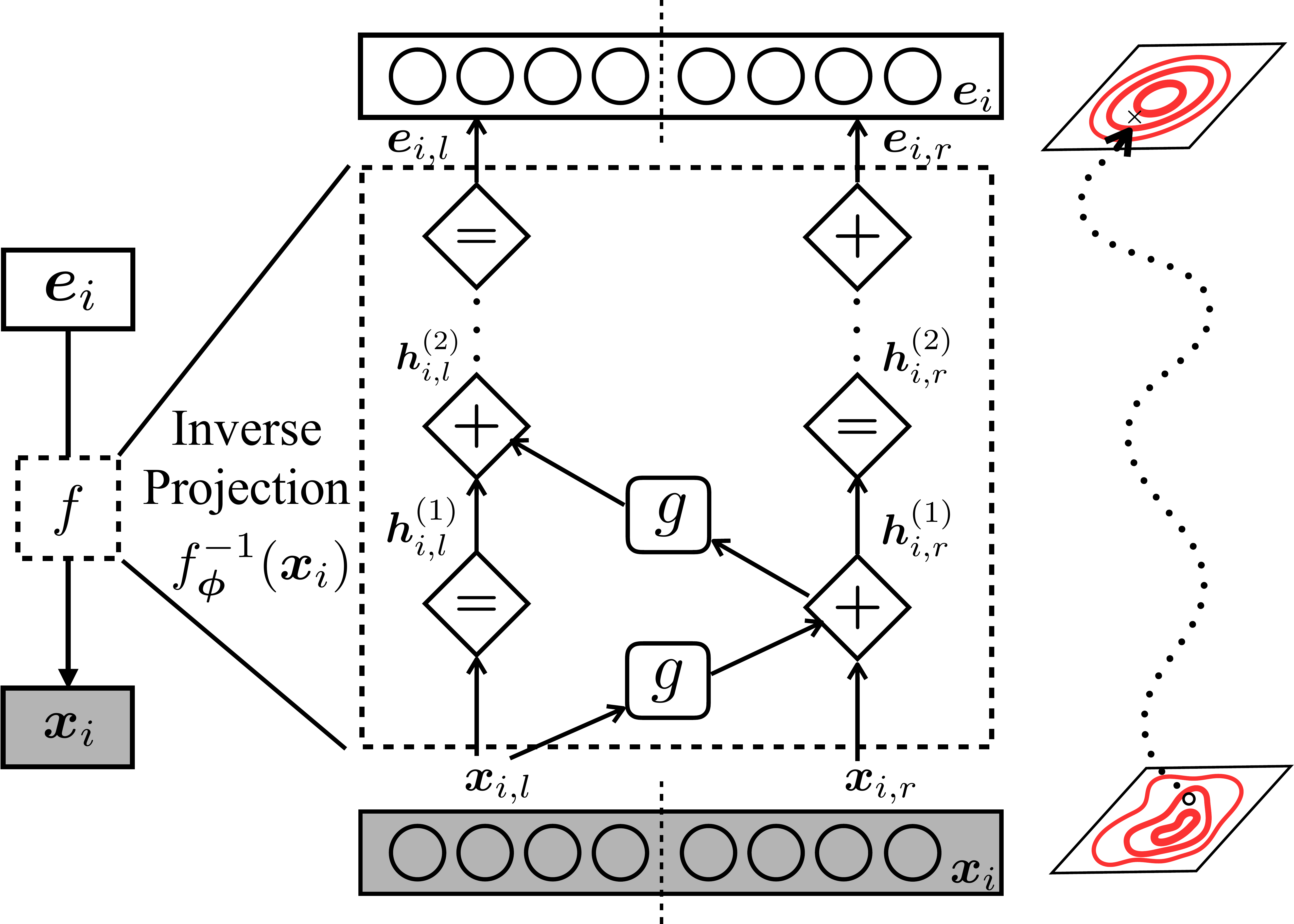}
    \caption{Depiction of the architecture of the inverse projection $f_{\bm{\phi}}^{-1}$ that composes multiple volume-preserving coupling layers, with which we parameterize our model. On the right, we schematically depict how the inverse projection transforms the observed word embedding $\bm{x}_i$ to a point $\bm{e}_i$ in a new embedding space.}
    \label{fig:opt}
\vspace{-5mm}
\end{figure}

From Eq.~\eqref{eq:learn-general} we know that only $f_{\bm{\phi}}^{-1}$ is required for accomplishing learning and inference; we never need to explicitly construct $f_{\bm{\phi}}$. Thus, we directly define the architecture of $f_{\bm{\phi}}^{-1}$. As shown in Figure \ref{fig:opt}, the nonlinear transformation from the observed embedding $\bm{x}_i$ to $\bm{h}_i^{(1)}$ represents the first coupling layer. The input in this layer is partitioned into left and right halves of dimensions, $\bm{x}_{i, l}$ and $\bm{x}_{i, r}$,  respectively. A single coupling layer is defined as:
\begin{equation}
\bm{h}_{i, l}^{(1)} = \bm{x}_{i, l}, \qquad
\bm{h}_{i, r}^{(1)} = \bm{x}_{i, r} + g(\bm{x}_{i, l}),
\end{equation}
where $g: \mathbb{R}^{d_x/2} \rightarrow \mathbb{R}^{d_x/2}$ is the coupling function and can be any nonlinear form. This transformation satisfies $\text{dim}(\bm{h}^{(1)}) = \text{dim}(\bm{x})$, and \citet{dinh2014nice} show that its Jacobian matrix is triangular with all ones on the main diagonal. Thus the Jacobian determinant is always equal to one (i.e.~volume-preserving) and the invertibility condition is naturally satisfied. 

To be sufficiently expressive, we compose multiple coupling layers as suggested in~\citet{dinh2014nice}. Specifically, we exchange the role of left and right half vectors at each layer as shown in Figure \ref{fig:opt}. For instance, from $\bm{x}_i$ to $\bm{h}_i^{(1)}$ the left subset $\bm{x}_{i, l}$ is unchanged, while from $\bm{h}_i^{(1)}$ to $\bm{h}_i^{(2)}$ the right subset $\bm{h}_{i, r}^{(1)}$ remains the same. Also note that composing multiple coupling layers does not change the volume-preserving and invertibility properties. Such a sequence of invertible transformations from the data space $\bm{x}$ to $\bm{e}$ is also called normalizing flow~\citep{rezende2015variational}.


\section{Experiments}
In this section, we first describe our datasets and experimental setup. We then instantiate our approach with Markov and DMV-structured syntax models, and report results on POS tagging and dependency grammar induction respectively. Lastly, we analyze the learned latent embeddings. 

\subsection{Data}
For both POS tagging and dependency parsing, we run experiments on the Wall Street Journal (WSJ) portion of the Penn Treebank.\footnote{Preprocessing is different for the two tasks, we describe the details in the following subsections.} To create the observed data embeddings, we train skip-gram word embeddings~\citep{mikolov2013efficient} that are found to capture syntactic properties well when trained with small context window~\citep{bansal2014tailoring, lin2015unsupervised}. Following \citet{lin2015unsupervised}, the dimensionality $d_x $ is set to 100, and the training context window size is set to 1 to encode more syntactic information. The skip-gram embeddings are trained on the one billion word language modeling benchmark dataset~\citep{chelba2013one} in addition to the WSJ corpus.

\subsection{General Experimental Setup}
For the neural projector, we employ rectified networks as coupling function $g$ following \citet{dinh2014nice}. We use a rectified network with an input layer, one hidden layer, and linear output units, the number of hidden units is set to the same as the number of input units. The number of coupling layers are varied as 4, 8, 16 for both tasks. We optimize marginal data likelihood directly using Adam~\citep{kingma2014adam}. For both tasks in the fully unsupervised setting, we do not tune the hyper-parameters using supervised data.

\subsection{Unsupervised POS tagging}
\label{sec:pos}

For unsupervised POS tagging, we use a Markov-structured syntax model in our approach, which is a popular structure for unsupervised tagging tasks~\citep{lin2015unsupervised,tran2016unsupervised}.

\paragraph{Setup.}
Following existing literature, we train and test on the entire WSJ corpus (49208 sentences, 1M tokens). We use 45 tag clusters, the number of POS tags that appear in WSJ corpus. We train the discrete HMM and the Gaussian HMM~\citep{lin2015unsupervised} as baselines. 
For the Gaussian HMM, mean vectors of Gaussian emissions are initialized with the empirical mean of all word vectors with an additive noise.
We assume diagonal covariance matrix for $p(\bm{e}_i|z_i)$ and initialize it with the empirical variance of the word vectors. Following \citet{lin2015unsupervised}, the covariance matrix is fixed during training. The multinomial probabilities are initialized as $\theta_{kv} \propto \exp (u_{kv})$, where $u_{kv} \sim U[0, 1]$.  For our approach, we initialize the syntax model and Gaussian parameters with the pre-trained Gaussian HMM. The weights of layers in the rectified network are initialized from a uniform distribution with mean zero and a standard deviation of $\sqrt{1/n_{\text{in}}}$, where $n_{in}$ is the input dimension.\footnote{This is the default parameter initialization in PyTorch.} We evaluate the performance of POS tagging with both Many-to-One (M-1) accuracy~\citep{johnson2007doesn} and V-Measure (VM)~\citep{rosenberg2007v}. Given a model we found that the tagging performance is well-correlated with the training data likelihood, thus we use training data likelihood as a unsupervised criterion to select the trained model over 10 random restarts after training 50 epochs. We repeat this process 5 times and report the mean and standard deviation of performance.

\begin{table}[!t]
	\centering
	\hspace{-0.25cm}
	\resizebox{0.97 \columnwidth}{!}{
	\begin{tabularx}{1.2\columnwidth}{@{}lc c}
	\toprule
	\textbf{System} & \textbf{M-1} & \textbf{VM} \\
	\hline
    \multicolumn{3}{c}{w/o hand-engineered features} \vspace{0.7mm}\\
    	Discrete HMM & 62.7 & 53.8\\
	PYP-HMM~{\scriptsize \citep{blunsom2011hierarchical}} & 77.5 & 69.8 \\
	NHMM~{\scriptsize(basic)~\citep{tran2016unsupervised}} & 59.8 & 54.2 \\
	NHMM~{\scriptsize(+ Conv)~\citep{tran2016unsupervised}} & 74.1 & 66.1 \\
	NHMM~{\scriptsize(+ Conv \& LSTM)~\citep{tran2016unsupervised}} & 79.1 & 71.7 \\ 
	Gaussian HMM~{\scriptsize \citep{lin2015unsupervised}} & 75.4 (1.0) & 68.5 (0.5) \\ \hdashline
        Ours (4 layers) & 79.5 (0.9) & 73.0 (0.7) \\
        Ours (8 layers) & \textbf{80.8} (1.3) & \textbf{74.1} (0.7) \\
        Ours (16 layers) & 73.2 (4.3) & 70.5 (2.1) \\
	\midrule
    \multicolumn{3}{c}{w/ hand-engineered features} \vspace{0.7mm}\\
    	Feature HMM~{\scriptsize \citep{berg2010painless}} & 75.5 & -- \\
	Brown~{\scriptsize(+ proto)~\citep{christodoulopoulos2010two}} & 76.1 & 68.8 \\ 
        Cluster~{\scriptsize(word-based)~\citep{yatbaz2012learning}} & 80.2 & 72.1 \\
        Cluster~{\scriptsize(token-based)~\citep{yatbaz2014unsupervised}} & 79.5 & 69.1 \\
	\bottomrule 
	\end{tabularx}}
	\caption{Unsupervised POS tagging results on entire WSJ, compared with other baselines and state-of-the-art systems. Standard deviation is given in parentheses when available.}
	\label{tab:pos-result}
	\vspace{-5mm}
\end{table}

\paragraph{Results. }
We compare our approach with basic HMM, Gaussian HMM, and several state-of-the-art systems,
including sophisticated HMM variants and clustering techniques with hand-engineered features. The results are presented in Table~\ref{tab:pos-result}. Through the introduced latent embeddings and additional neural projection, our approach improves over the Gaussian HMM by 5.4 points in M-1 and 5.6 points in VM. Neural HMM (NHMM)~\citep{tran2016unsupervised} is a baseline that also learns word representation jointly. Both their basic model and extended Conv version does not outperform the Gaussian HMM. Their best model incorporates another LSTM to model long distance dependency and breaks the Markov assumption, yet our approach still achieves substantial improvement over it without considering more context information. Moreover, our method outperforms the best published result that benefits from hand-engineered features~\citep{yatbaz2012learning} by 2.0 points on VM.   

\begin{figure}[!t]
 \begin{center}
 \subfigure[Gaussian HMM]{
 \includegraphics[scale=0.295]{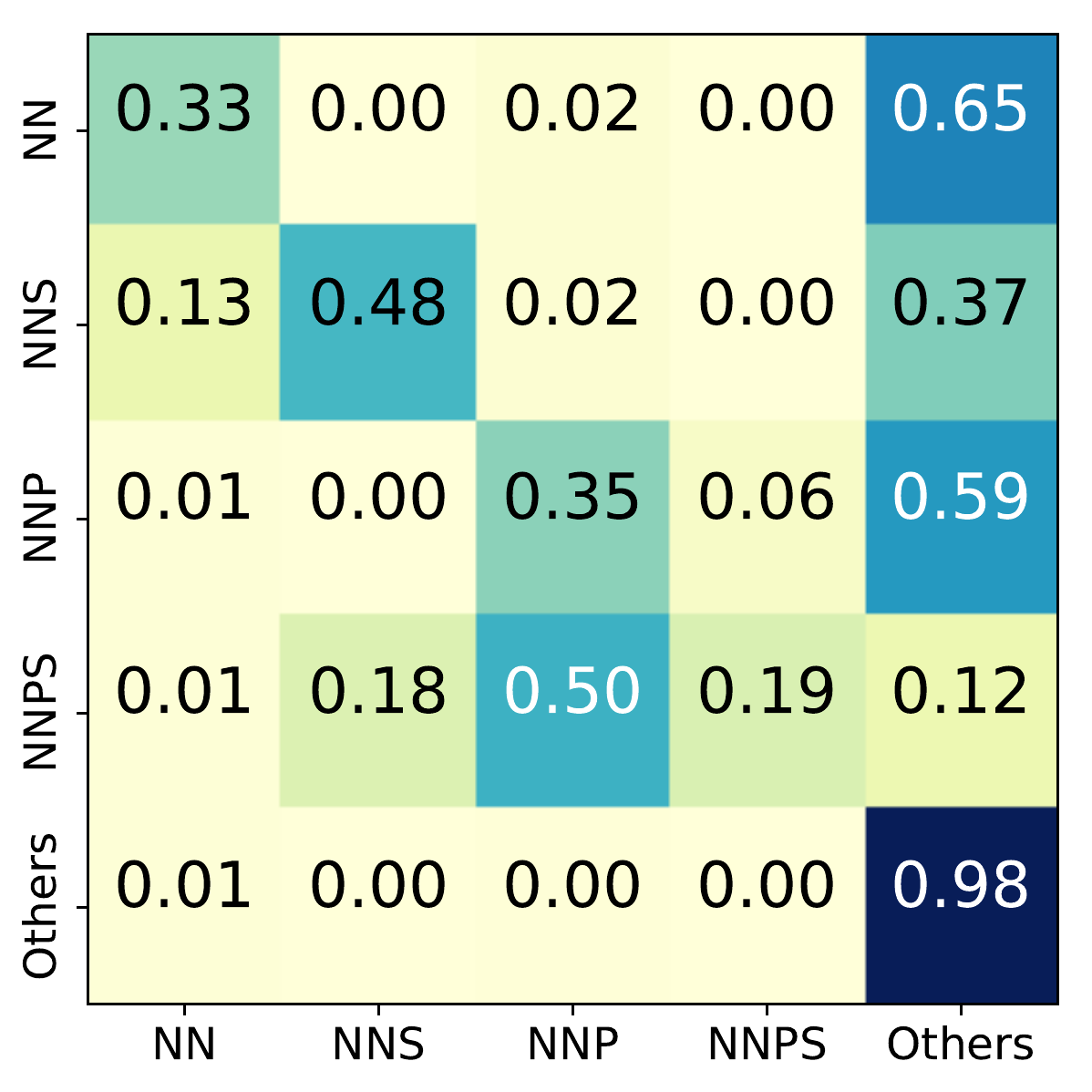}
 }
 \hfil
 \subfigure[Our approach]{
 \includegraphics[scale=0.295]{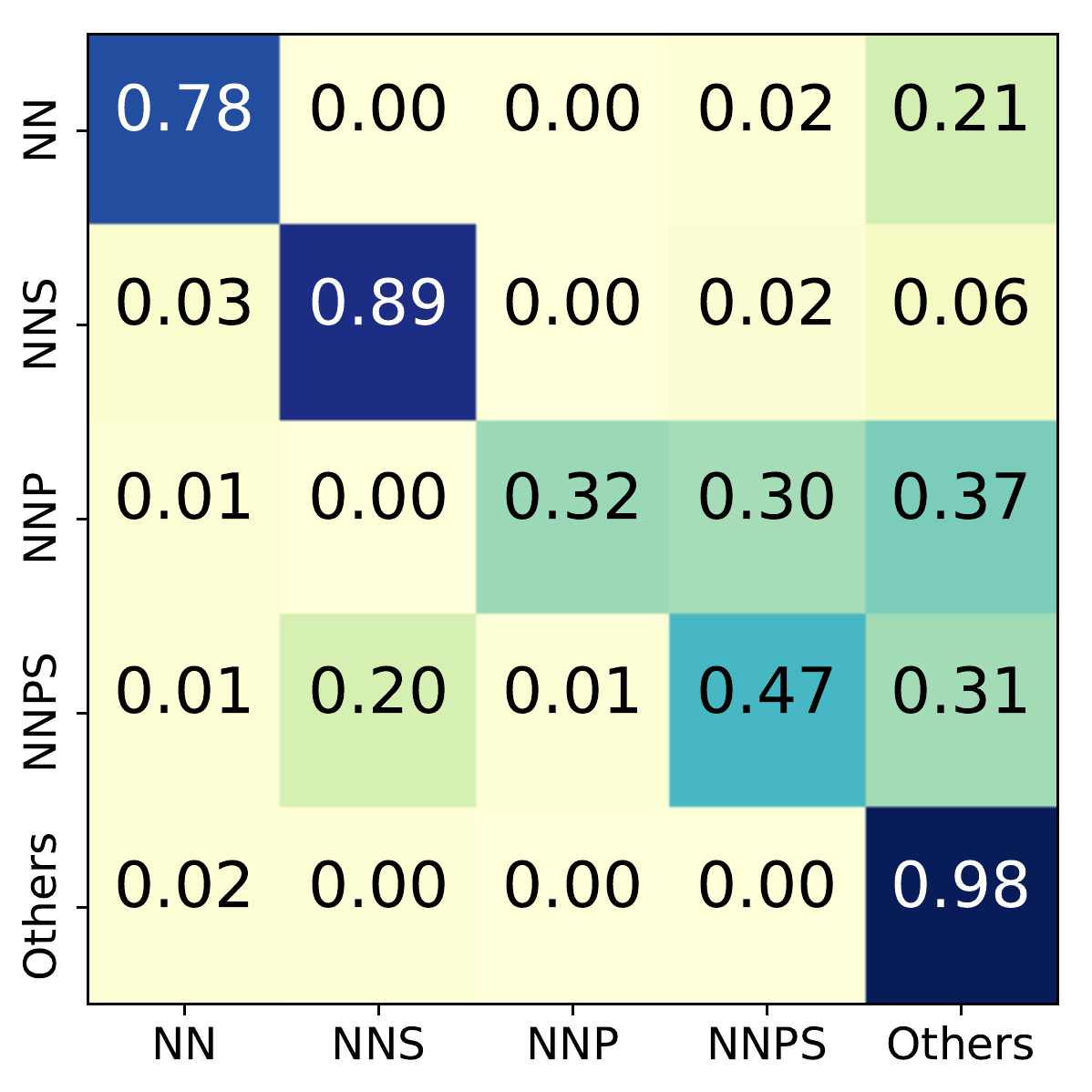}
 }
 \vspace{-3mm}
 \caption{Normalized Confusion matrix for POS tagging experiments, row label represents the gold tag.}
  \vspace{-5mm}
 \label{fig:cm}
 \end{center}
 \end{figure}

 \paragraph{Confusion Matrix. }
We found that most tagging errors happen in noun subcategories. Therefore, we do the one-to-one mapping between gold POS tags and induced clusters and plot the normalized confusion matrix of noun subcategories in Figure~\ref{fig:cm}. The Gaussian HMM fails to identify ``NN'' and ``NNS'' correctly for most cases, and it often recognizes ``NNPS'' as ``NNP''. In contrast, our approach corrects these errors well.

\subsection{Unsupervised Dependency Parsing without gold POS tags}
For the task of unsupervised dependency parse induction, we employ the Dependency Model with Valence (DMV)~\citep{klein2004corpus} as the syntax model in our approach. DMV is a generative model that defines a probability distribution over dependency parse trees and syntactic categories, generating tokens and dependencies in a head-outward fashion. 
While, traditionally, DMV is trained using gold POS tags as observed syntactic categories, in our approach, we treat each tag as a latent variable, as described in \cref{sec:general-neural}. 

Most existing approaches to this task are not fully unsupervised since they rely on gold POS tags following the original experimental setup for DMV. This is partially because automatically parsing from words is difficult even when using unsupervised syntactic categories~\citep{spitkovsky2011unsupervised}. However, inducing dependencies from words alone represents a more realistic experimental condition since gold POS tags are often unavailable in practice. Previous work that has trained from words alone often requires additional linguistic constraints (like sentence internal boundaries)~\citep{spitkovsky2011unsupervised, spitkovsky2011punctuation, spitkovsky2012capitalization, spitkovsky2013breaking}, acoustic cues~\citep{pate2013unsupervised}, additional training data~\citep{pate2016grammar}, or annotated data from related languages~\citep{cohen2011unsupervised}. Our approach is naturally designed to train on word embeddings directly, thus we attempt to induce dependencies without using gold POS tags or other extra linguistic information.

\begin{table}[!t]
	\centering
	\hspace{-0.25cm}
	\resizebox{0.97 \columnwidth}{!}{
	\begin{tabularx}{1.2 \columnwidth}{@{}lc c}
	\toprule
	\textbf{System} & $\leqslant 10$ & all\\
	\hline
    \multicolumn{3}{c}{w/o gold POS tags} \vspace{0.7mm}\\
    	DMV {\scriptsize \citep{klein2004corpus}} & 49.6 & 35.8 \\
        E-DMV {\scriptsize \citep{headden2009improving}} & 52.1 & 38.2 \\
        UR-A E-DMV~{\scriptsize \citep{tu2012unambiguity}} & 58.9 & 46.1 \\
        CS$^{\ast}$ {\scriptsize \citep{spitkovsky2013breaking}} & \hspace{2pt} 72.0$^{\ast}$ & \hspace{2pt} 64.4$^{\ast}$ \\
        Neural E-DMV {\scriptsize \citep{jiang2016unsupervised}} & 55.3 & 42.7\\
	CRFAE {\scriptsize \citep{cai2017crf}} & 37.2 & 29.5 \\ 
	Gaussian DMV & 55.4 (1.3) & 43.1 (1.2) \\ \hdashline
        Ours (4 layers) & 58.4 (1.9) & 46.2 (2.3) \\
        Ours (8 layers) & \textbf{60.2} (1.3) & \textbf{47.9} (1.2) \\
        Ours (16 layers) & 54.1 (8.5) & 43.9 (5.7) \\
	\midrule
    \multicolumn{3}{c}{w/ gold POS tags {\small (for reference only)}} \vspace{0.7mm}\\
        DMV~{\scriptsize \citep{klein2004corpus}} & 55.1 & 39.7 \\
        UR-A E-DMV~{\scriptsize \citep{tu2012unambiguity}} & 71.4 & 57.0 \\
        MaxEnc~{\scriptsize \citep{le2015unsupervised}} & 73.2 & 65.8 \\
        Neural E-DMV~{\scriptsize \citep{jiang2016unsupervised}} & 72.5 & 57.6 \\
        CRFAE~{\scriptsize \citep{cai2017crf}} & 71.7 & 55.7 \\
        L-NDMV~{\scriptsize (Big training data)~\citep{han2017dependency}} & 77.2 & 63.2 \\
	\bottomrule 
	\end{tabularx}}
	\caption{Directed dependency accuracy on section 23 of WSJ, evaluating on sentences of length $\leqslant 10$ and all lengths. Starred entries ($\ast$) denote that the system benefits from additional punctuation-based constraints. Standard deviation is given in parentheses when available.}
	\label{tab:dmv-result}
 	\vspace{-7mm}
\end{table}

\paragraph{Setup.}
Like previous work we use sections 02-21 of WSJ corpus as training data and evaluate on section 23,  we remove punctuations and train the models on sentences of length $\leqslant 10$, ``head-percolation'' rules~\citep{collins1999head} are applied to obtain gold dependencies for evaluation. We train basic DMV, extended DMV (E-DMV)~\citep{headden2009improving} and Gaussian DMV (which treats POS tag as unknown latent variables and generates observed word embeddings directly conditioned on them following Gaussian distribution) as baselines. Basic DMV and E-DMV are trained with Viterbi EM~\citep{spitkovsky2010viterbi} on unsupervised POS tags induced from our Markov-structured model described in \cref{sec:pos}. Multinomial parameters of the syntax model in both Gaussian DMV and our model are initialized with the pre-trained DMV baseline. Other parameters are initialized in the same way as in the POS tagging experiment. The directed dependency accuracy (DDA) is used for evaluation and we report accuracy on sentences of length $\leqslant 10$ and all lengths. We train the parser until training data likelihood converges, and report the mean and standard deviation over 20 random restarts. 

\paragraph{Comparison with other related work.}
Our model directly observes word embeddings and does not require gold POS tags during training. Thus, results from related work trained on gold tags are not directly comparable. However, to measure how these systems might perform without gold tags, we run three recent state-of-the-art systems in our experimental setting: UR-A E-DMV~\citep{tu2012unambiguity}, Neural E-DMV~\citep{jiang2016unsupervised}, and CRF Autoencoder (CRFAE)~\citep{cai2017crf}.\footnote{For the three systems, we use implementations from the original papers (via personal correspondence with the authors), and tune their hyperparameters on section 22 of WSJ.} We use unsupervised POS tags (induced from our Markov-structured model) in place of gold tags.\footnote{Using words directly is not practical because these systems often require a transition probability matrix between input symbols, which requires too much memory.} We also train basic DMV on gold tags and include several state-of-the-art results on gold tags as reference points.

\paragraph{Results.}
As shown in Table \ref{tab:dmv-result}, our approach is able to improve over the Gaussian DMV by 4.8 points on length $\leqslant 10$ and 4.8 points on all lengths, which suggests the additional latent embedding layer and neural projector are helpful. The proposed approach yields, to the best of our knowledge,\footnote{We tried to be as thorough as possible in evaluation by running top performing systems using our more difficult training setup when this was feasible -- but it was not possible to evaluate them all.} state-of-the-art performance without gold POS annotation and without sentence-internal boundary information. DMV, UR-A E-DMV, Neural E-DMV, and CRFAE suffer a large decrease in performance when trained on unsupervised tags -- an effect also seen in previous work~\citep{spitkovsky2011unsupervised, cohen2011unsupervised}. Since our approach induces latent POS tags jointly with dependency trees, it may be able to learn POS clusters that are more amenable to grammar induction than the unsupervised tags. We observe that CRFAE underperforms its gold-tag counterpart substantially. This may largely be a result of the model's reliance on prior linguistic rules that become unavailable when gold POS tag types are unknown. 
Many extensions to DMV can be considered orthogonal to our approach -- they essentially focus on improving the syntax model.
It is possible that incorporating these more sophisticated syntax models into our approach may lead to further improvements.

\subsection{Sensitivity Analysis}
\paragraph{Impact of Initialization. }
\begin{table}[!t]
    \centering
    \resizebox{0.97 \columnwidth}{!}{
    \begin{tabularx}{1.2 \columnwidth}{l@{\hskip 1.5cm}c@{\hskip 2.5cm}c}
    \toprule
      \textbf{System}   & \textbf{M-1} & \textbf{VM} \\
    \hline
       Ours (4 layers) & 78.2 & 71.2 \\
       Ours (8 layers) & 72.5 & 69.7 \\
       Ours (16 layers) & 67.2 & 69.2 \\
    \bottomrule
    \end{tabularx}}
    \caption{Unsupervised POS tagging results of our approach on WSJ, with random initialization of syntax model.}
    \label{tab:pos-rand-result}
    \vspace{-5mm}
\end{table}
In the above experiments we initialize the structured syntax components with the pre-trained Gaussian or discrete baseline, which is shown as a useful technique to help train our deep models. We further study the results with fully random initialization. In the POS tagging experiment, we report the results in Table~\ref{tab:pos-rand-result}. While the performance with 4 layers is comparable to the pre-trained Gaussian initialization, deeper projections (8 or 16 layers) result in a dramatic drop in performance. This suggests that the structured syntax model with very deep projections is difficult to train from scratch, and a simpler projection might be a good compromise in the random initialization setting.   

Different from the Markov prior in POS tagging experiments, our parsing model seems to be quite sensitive to the initialization. For example, directed accuracy of our approach on sentences of length $\leqslant 10$ is below 40.0 with random initialization. This is consistent with previous work that has noted the importance of careful initialization for DMV-based models such as the commonly used harmonic initializer~\citep{klein2004corpus}. However, it is not straightforward to apply the harmonic initializer for DMV directly in our model without using some kind of pre-training since we do not observe gold POS.

\paragraph{Impact of Observed Embeddings. }
\begin{table}[!t]
    \centering
    \resizebox{0.97 \columnwidth}{!}{
    \begin{tabularx}{1.2 \columnwidth}{l@{\hskip 1.5cm}c@{\hskip 2.5cm}c}
    \toprule
      \textbf{System}   & \textbf{M-1} & \textbf{VM} \\
    \hline
    Gaussian HMM & 72.0 & 65.0\\
    Ours (4 layers) & 76.4 & 69.3 \\
    Ours (8 layers) & 76.8 & 69.4 \\
    Ours (16 layers) & 67.3 & 62.0 \\
    \bottomrule
    \end{tabularx}}
    \caption{Unsupervised POS tagging results on WSJ, with fastText vectors as the observed embeddings.}
    \label{tab:pos-fasttext-result}
\end{table}

\begin{table}[!t]
    \centering
    \resizebox{0.97 \columnwidth}{!}{
    \begin{tabularx}{1.2 \columnwidth}{l@{\hskip 1.5cm}c@{\hskip 2.5cm}c}
    \toprule
      \textbf{System} &  $\leqslant 10$ & all \\
    \hline
    Gaussian DMV & 53.6 & 41.3\\
    Ours (4 layers) & 56.9 & 43.9 \\
    Ours (8 layers) & 57.1 & 42.3 \\
    Ours (16 layers) & 52.9 & 39.5 \\
    \bottomrule
    \end{tabularx}}
    \caption{Directed dependency accuracy on section 23 of WSJ, with fastText vectors as the observed embeddings.}
    \label{tab:parse-fasttext-result}
    \vspace{-5mm}
\end{table}
We investigate the effect of the choice of pre-trained embedding on performance while using our approach. To this end, we additionally include results using fastText embeddings~\citep{bojanowski2017enriching} -- which, in contrast with skip-gram embeddings, include character-level information. We set the context windows size to 1 and the dimension size to 100 as in the skip-gram training, while keeping other parameters set to their defaults. These results are summarized in Table~\ref{tab:pos-fasttext-result} and Table~\ref{tab:parse-fasttext-result}. While fastText embeddings lead to reduced performance with our model, our approach still yields an improvement over the Gaussian baseline with the new observed embeddings space.

\subsection{Qualitative Analysis of Embeddings}

We perform qualitative analysis to understand how the latent embeddings help induce syntactic structures. First we filter out low-frequency words and punctuations in WSJ, and visualize the rest words (10k) with t-SNE~\citep{maaten2008visualizing} under different embeddings. We assign each word with its most likely gold POS tags in WSJ and color them according to the gold POS tags.

\begin{table}[!t]
	\centering
	\small
	\resizebox{1.0 \columnwidth}{!}{
	\begin{tabularx}{\columnwidth}{X | p{2.8cm} | p{2.6cm} l}
	\hline
	\textbf{Target} & \textbf{Skip-gram} & \textbf{Markov Structure} \\
	\hline
    	come & go came follow \newline coming sit& be go do give \newline follow\\
    \hline
   singing & dancing sing \newline drumming dance \newline dances &  dancing drumming \newline marching playing \newline recording \\
\hline
        cigars & cigarettes sodas \newline champagne cigar \newline rum & sodas bottles \newline drinks pills \newline cigarettes\\
    \hline
       newer & flashier fancier \newline conventional low-end \newline new-generation & softer lighter \newline thinner darker \newline smoother\\
    \hline
      fanciest & priciest up-scale \newline loveliest fancier \newline high-end & liveliest priciest \newline smartest best-run \newline fastest-growing\\
	\hline
	\end{tabularx}}
	\caption{Target words and their 5 nearest neighbors,  based on skip-gram embeddings and our learned latent embeddings with Markov-structured syntax model.}
	\label{tab:emb-case}
	\vspace{-3mm}
\end{table}

For our Markov-structured model, we have displayed the embedding space in Figure~\ref{fig:vi-pos-mar}, where the gold POS clusters are well-formed. Further, we present five example target words and their five nearest neighbors in terms of cosine similarity. As shown in  Table~\ref{tab:emb-case}, the skip-gram embedding captures both semantic and syntactic aspects to some degree, yet our embeddings are able to focus especially on the syntactic aspects of words, in an unsupervised fashion without using any extra morphological information. 

\begin{figure}[t]
  \centering
   \includegraphics[scale=0.23]{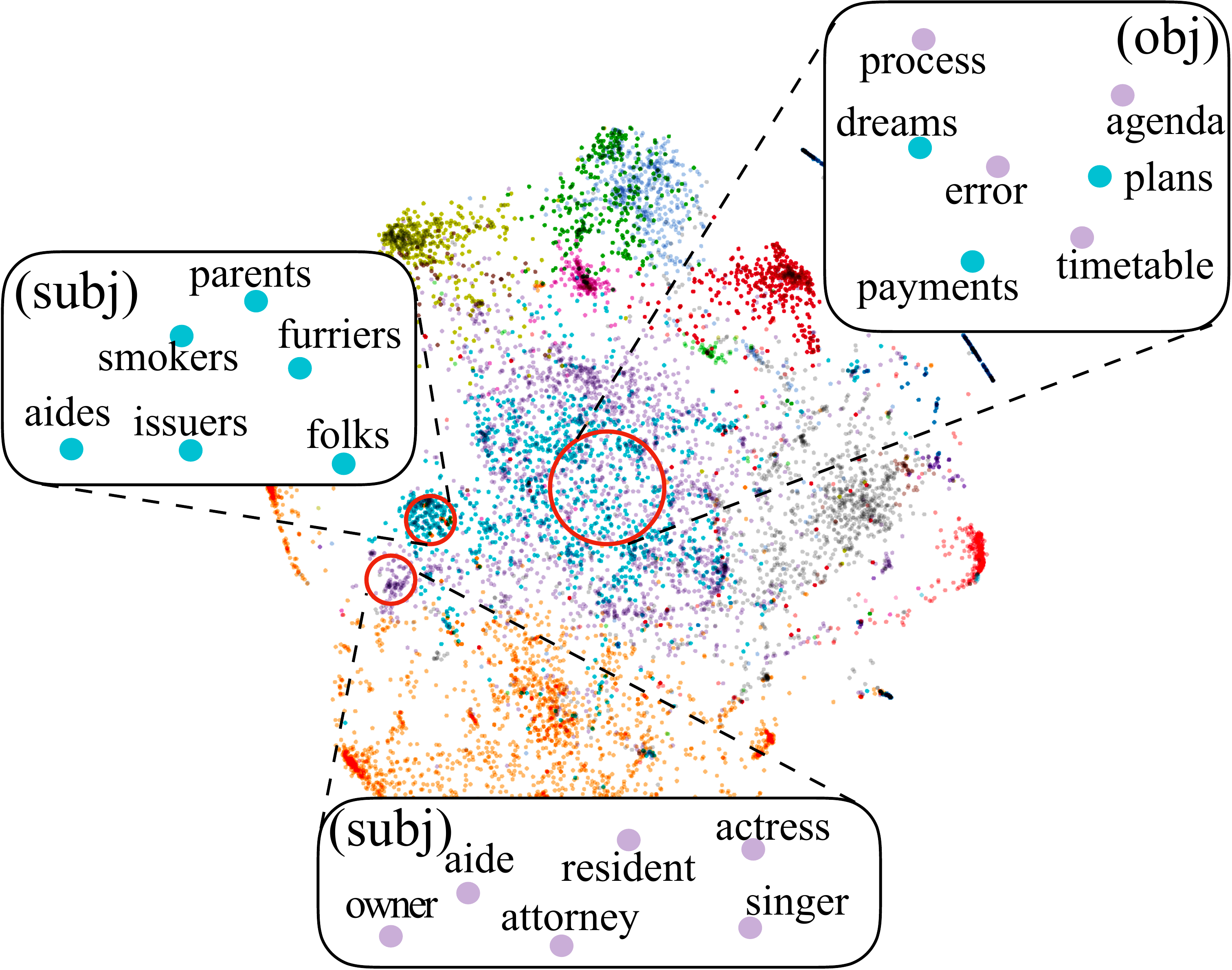}
   \vspace{-2mm}
   \caption{Visualization (t-SNE) of learned latent embeddings with DMV-structured syntax model. Each node represents a word and is colored according to the most likely gold POS tag in the Penn Treebank (best seen in color).}
   \label{fig:vi-dmv}
   \vspace{-5mm}
\end{figure}

In Figure~\ref{fig:vi-dmv} we depict the learned latent embeddings with the DMV-structured syntax model. Unlike the Markov structure, the DMV structure maps a large subset of singular and plural nouns to the same overlapping region. However, two clusters of singular and plural nouns are actually separated. We inspect the two clusters and the overlapping region in Figure~\ref{fig:vi-dmv}, it turns out that the nouns in the separated clusters are words that can appear as subjects and, therefore, for which verb agreement is important to model. In contrast, the nouns in the overlapping region are typically objects. This demonstrates that the latent embeddings are focusing on aspects of language that are specifically important for modeling dependency without ever having seen examples of dependency parses. 

Some previous work has deliberately created embeddings to capture different notions of similarity~\citep{levy2014dependency, cotterell2015morphological}, while they use extra morphology or dependency annotations to guide the embedding learning, our approach provides a potential alternative to create new embeddings that are guided by structured syntax model, only using unlabeled text corpora.
\section{Related Work}


Our approach is related to flow-based generative models, which are first described in NICE~\citep{dinh2014nice} and have recently received more attention~\citep{dinh2016density, jacobsen2018revnet, kingma2018glow}. This relevant work mostly adopts simple (e.g. Gaussian) and fixed priors and does not attempt to learn interpretable latent structures. Another related generative model class is variational auto-encoders (VAEs)~\citep{kingma2013auto} that optimize a lower bound on the marginal data likelihood, and can be extended to learn latent structures~\citep{miao2016language, yin2018structvae}. Against the flow-based models, VAEs remove the invertibility constraint but sacrifice the merits of exact inference and exact log likelihood computation, which potentially results in optimization challenges~\citep{kingma2016improved}. 
Our approach can also be viewed in connection with generative adversarial networks (GANs)~\citep{goodfellow2014generative} that is a likelihood-free framework to learn implicit generative models. However, it is non-trivial for a gradient-based method like GANs to propagate gradients through discrete structures.

\section{Conclusion}

In this work, we define a novel generative approach to leverage continuous word representations for unsupervised learning of syntactic structure. Experiments on both POS induction and unsupervised dependency parsing tasks demonstrate the effectiveness of our proposed approach. Future work might explore more sophisticated invertible projections, or recurrent projections that jointly transform the entire input sequence. 
\bibliography{emnlp2018}
\bibliographystyle{acl_natbib_nourl}

\end{document}